\documentclass[letterpaper, 10 pt, conference]{ieeeconf}  

\IEEEoverridecommandlockouts                              

\usepackage{color}
\usepackage[normalem]{ulem}
\usepackage{graphicx}
\usepackage{multirow}
\usepackage[export]{adjustbox}
\usepackage{caption}
\usepackage{subfig}

\definecolor{mypink2}{rgb}{0.8, 0.35, 0.31}
\definecolor{mypink1}{rgb}{0, 0, 0}
\definecolor{erhcomc}{rgb}{0, 0.7, 0.0}
\definecolor{soutc}{rgb}{0.5, 0.5, 0.5}
\definecolor{chocolate}{rgb}{0,0,0}
\definecolor{purple}{rgb}{0.5,0.0,0.5}
\definecolor{blue}{rgb}{0.05, 0.05, 0.9}
\definecolor{green}{rgb}{0.25, 0.9, 0.7}

\newcommand{\tuba}[1]{\textcolor{chocolate}{#1}}

\newcommand\Initial{pre-MP}
\newcommand\Correction{cor-MP}
\newcommand\Final{post-MP}

\newcommand\InitialCNMP{pre-MP network}
\newcommand\CorrectionCNMP{cor-MP network}
\newcommand\FinalCNMP{post-MP network}

\newcommand\AECNMP{AE-CNMPs}
\newcommand\PCCNMP{PC-CNMPs}
\newcommand\CNMP{CNMPs}
\newcommand\Baseline{Vanilla CNMPs}

\title{\LARGE \bf
Bimanual rope manipulation skill synthesis through context dependent correction policy learning from human demonstration

}

\author{T. Baturhan Akbulut$^{1,*}$, G. Tuba C. Girgin$^{1,*}$, Arash Mehrabi$^{2}$,  Minoru Asada$^{3,4}$, Emre Ugur$^{1}$, Erhan Oztop$^{2,3}$
\thanks{$^{*}$ Equal contribution, alphabetical order}
\thanks{$^{1}$ T. Baturhan Akbulut, G. Tuba C. Girgin and Emre Ugur are with Deparment of Computer Engineering,
        Bogazici University, Istanbul, Turkey. $^{2}$ Arash Mehrabi and Erhan Oztop are with Department of Computer Engineering, Ozyegin University, Istanbul, Turkey. $^{3}$ Erhan Oztop and Minoru Asada are with Symbiotic Intelligent Systems Research Center, Institute for Open and Transdisciplinary, Research Initiatives, Osaka University, Japan. $^{4}$ Minoru Asada is also affiliated with International Professional University of Technology in Osaka, Osaka, Japan.}%
}

\begin{document}

\maketitle
\thispagestyle{empty}
\pagestyle{empty}

\begin{abstract}

Learning from demonstration (LfD) provides a convenient means to equip robots with dexterous skills when demonstration can be obtained in robot intrinsic coordinates. However, the problem of compounding errors in long and complex skills reduces its wide deployment. Since most such complex skills are composed of smaller movements  that are combined, considering the target skill as a sequence of compact motor primitives seems reasonable. Here the problem that needs to be tackled is to ensure that a motor primitive ends in a state that allows the successful execution of the subsequent primitive. In this study, we focus on this problem by proposing to learn an explicit correction policy when the expected transition state between primitives is not achieved. The correction policy is itself learned via behavior cloning by the use of a state-of-the-art movement primitive learning architecture, Conditional Neural Motor Primitives (CNMPs). The learned  correction policy is then able to produce diverse movement trajectories in a context dependent way.  The advantage of the proposed system over learning the complete task as a single action is shown with a table-top setup in simulation, where an object has to be pushed through a corridor in two steps. Then, the applicability of the proposed method to bi-manual knotting in the real world is shown by equipping an upper-body humanoid robot with the skill of making knots over a bar in 3D space. The experiments show that the robot can perform successful knotting even when the faced correction cases are not part of the human demonstration set.


\end{abstract}

\section{INTRODUCTION}

\begin{figure}
\centering
\includegraphics[width=0.3\textwidth]{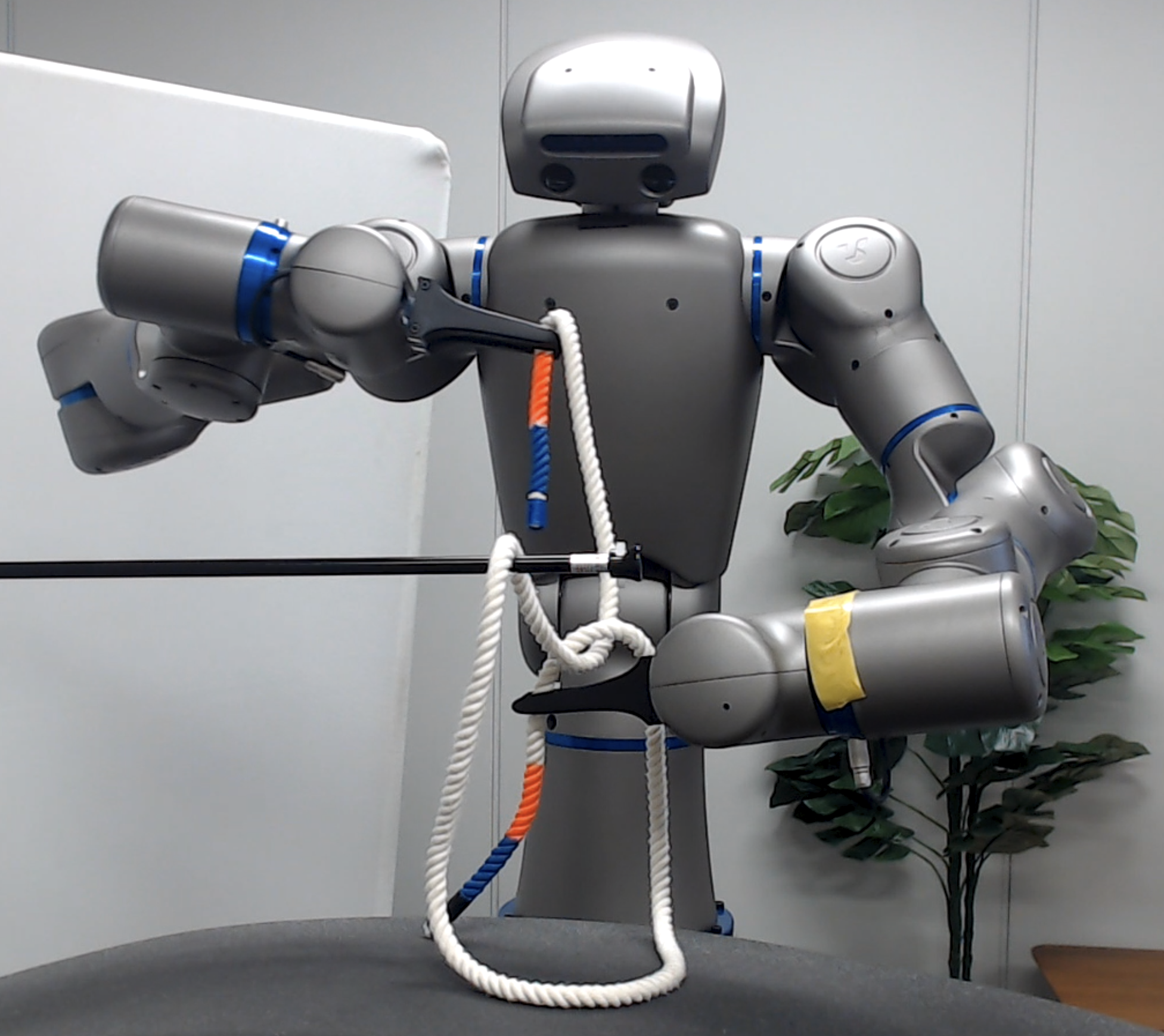}

\caption{\label{fig:real robot} Bi-manual knotting with Torobo in the real world.
}
\vspace{-8mm}
\end{figure}

Learning from demonstration (LfD) \cite{schaal1996learning} is an effective robot learning framework, where a robot is taught a desired skill through human demonstration. The demonstration may be kinaesthetic, visual, or through teleoperation \cite{amirshirzad-shared-19}. The latter places the human operator in the control loop. Thus human motor learning capacity enables teaching more complex tasks to the robot \cite{BabicHaleOztop2011}. The learning can be based on policy cloning \cite{pomerleau-beh-clon-88}, i.e. learning a state-to-action mapping, or simple motor-tape approach \cite{atkeson-hum-beh-2000}, i.e. playing back the stored motor commands in an open-loop fashion. For complex and long tasks, both methods suffer from the so called covariate shift problem \cite{chang-neurips-21}. When the system enters an off-learned distribution state, it can not recover. Even worse, it can diverge to states that are even further away. Assume a robot executing a rather long trajectory in order to make a knot using its two grippers. Even small mistakes the are made during the initial parts of the action such as holding the rope from a slightly different position or making slight errors in the movement trajectory that cause the gripper to miss the rope, would result in failure of task execution. A remedy is to consider the actions as the composition of simpler action primitives that are less prone to covariate  shift, allowing corrective actions to be inserted between primitives to ensure the successful execution of the subsequent primitive. Corrective action policies can be learned by the robot itself or again taught by a human demonstrator.  In this study, we explore the latter approach, first in simulation, then show that it can be applied to real hardware by synthesizing a complex rope manipulation skill on an upper-body humanoid robot (Fig~\ref{fig:real robot}).

In summary, we propose a method that enables the robot to recover from failure states that might be encountered during critical points in complex and long action executions. To do this, the action is segmented based on a set of identified critical points. During these critical points, the human operator demonstrates the movements that needs to be taken in order to make the task execution succeed, and the demonstration is learned by our system as a function of robot-environment context. The corrective actions that might require very different trajectories  are encoded in a single corrective movement primitive. Therefore, when encountered with a failure context, a corrective action is  automatically generated to ensure successful task execution.


Overall, the main contributions of our paper are (1) reducing the human effort for data generation for LfD by focusing the learning on correction policies, (2) showing that the approach is general by realizing it on real hardware, and (3) providing dexterous rope manipulation in 3D space superseding earlier rope knotting skills with real robots.

\section{Related Work}
There has been a significant amount of research on robotic manipulation of deformable objects. Nair et al. \cite{nair_rope_2017} combined self-supervised learning with imitation for rope manipulation. First, they trained an inverse dynamics model that predicts the action given the images of the current and next states. Then, they used human demonstration as a high-level plan of what to do and the inverse dynamics model to determine how to do it. The training data, which consisted of 60K pairs of images, were used for training the inverse dynamics model. 
The inverse dynamics model of \cite{nair_rope_2017} works well in predicting the actions that make small deformations, however it lacks the ability to construct a trajectory to reach to goal states. Wang et al. \cite{wang_visual_2019} extended the ideas of Nair et al. \cite{nair_rope_2017} by replacing human demonstration with a Casual InfoGAN (CIGAN) model. The CIGAN can generate a trajectory to reach a given goal state from the initial state. 
However, high success rates could not be reached in generating plausible plans, as explained in \cite{wang_visual_2019}.

Yan et al. \cite{yan_contrastive_2020} developed a visual model-based learning framework that jointly learns the underlying visual latent representations and the dynamics models for deformable objects. The latent dynamics model was learned in a simulation environment, then transferred to a real PR2 robot. In another work, Yan et al. \cite{yan_state_2019} proposed a model that learns the dynamics of linear deformable objects in state space. However, bi-manual rope manipulation was not addressed in either of the works, and they did not consider local corrective movements targeted at critical points as in our work.

Chang et al. \cite{chang-neurips-21} proposed the network MILO where offline RL achieves learning without needing online demonstrations. This method suffers from out-of-distribution configurations that a robot can encounter in real-life experiments. Reichlin et al. \cite{reichlin-recovery-2022} suggested a network to combine it with the offline RL setup to overcome the distribution-shift problem. The proposed recovery network encodes observations and creates a latent representation. The agent corrects its motion by using the encoded information and approximating the training dataset's density in latent space. However, the recovery network affects the predicted motion in each step, while our proposed method only needs correction in critical points, simplifying its logic.

Multiple other approaches have been tried for deformable object manipulation. Notably \cite{Lui-untagle-2013} realized a model-free visual reinforcement learning algorithm for fabric and rope manipulation both in simulated and real robot environments, and a max-margin learning algorithm for rope untangling demonstrated in the physical environment. 
Sundaresan et al. \cite{Sundaresan-DOD-2020} used an interpretable deep visual representations for rope. They learned point-pair correspondences between initial and goal rope configurations in simulation from synthetic depth data, enabling them to plan actions to reach the goal configuration, and applied their method also on a real robot. In another study, Seita et al. \cite{seita_imitation_2019} used deep imitation learning for fabric smoothing. They trained policies in simulation using dataset aggregation (DAgger) with an algorithmic supervisor, given images of a fabric sample as input. The difference between our algorithm and DAgger is that our algorithm only asks the expert for the correct action when it reaches a critical state, while DAgger needs continuous access to the supervisor's policy.


Unlike previous works, our algorithm learns to recover from failure states. We tackle the problem of learning an explicit correction policy using expert demonstration when the robot encounters a state where executing subsequent primitive is not possible.

\begin{figure}[t]
    
\centering
\includegraphics[width=0.4\textwidth]{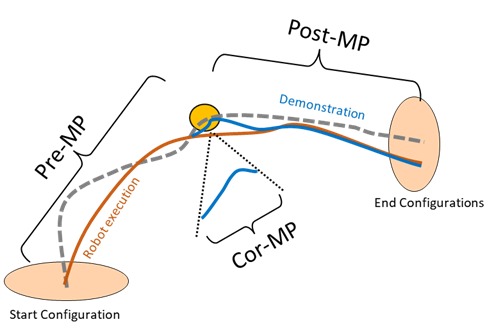}
\caption{\label{fig:methodimplementation}Illustration of the proposed concept with a single critical point. Demonstrated trajectory (dashed gray curve) passes through a critical region C, which depends on the task. For the robot execution, a correction movement is inserted to counteract the possible case of execution missing C, thereby splitting the execution into three motor primitives (MP): \Initial{}, \Correction{}, and \Final{}. \Final{} must be learned context-dependent to recover a wide range of failure cases. This is achieved by obtaining multiple correction demonstrations through teleoperation.}

\vspace{-5mm}
  
\end{figure}

\section{Method}
\label{methods}


\begin{figure*}%
    \centering
    \includegraphics[width=0.74\textwidth]{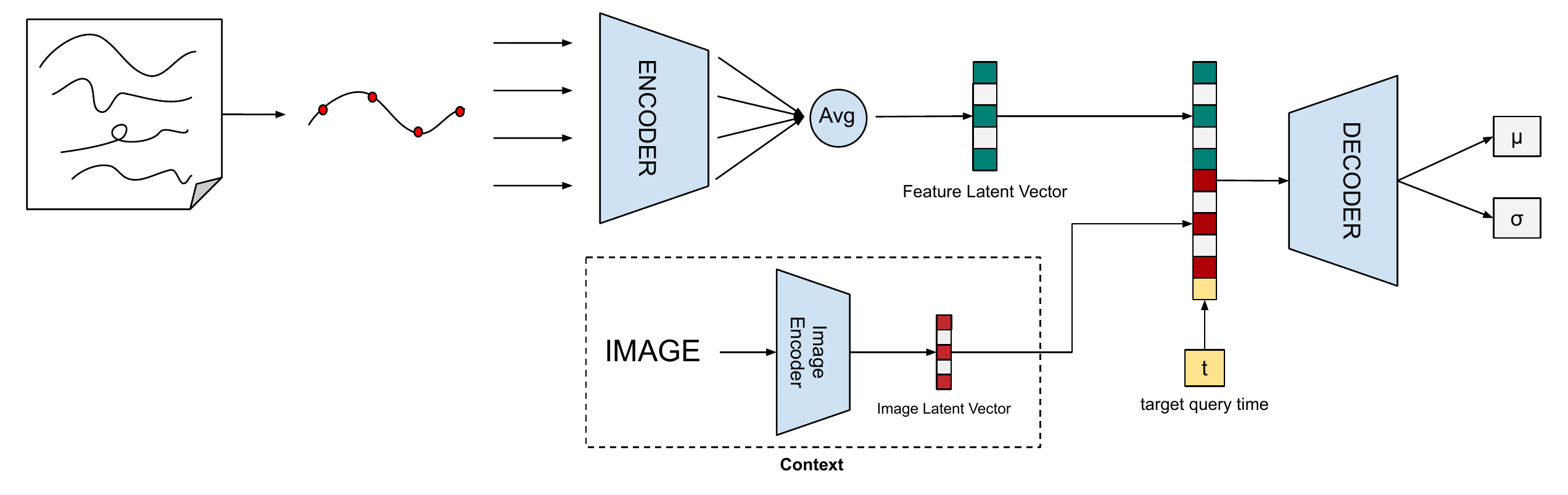}
    \caption{Conditional Neural Movement Primitive architecture (CNMP) with a context input is shown. In the illustration, the context is provided by an autoencoder. See Section~\ref{methods} for more details.}%
    \label{fig:cnmpwithcontextgraph}%
    
    \vspace{-5mm}
\end{figure*}

Complex tasks, e.g., rope manipulation, are often linked through critical points significant for successfully executing the task, where critical points are considered to be certain parts of the execution trajectory that may lead to failure.
Furthermore, long and complex tasks are fragile to the accumulation of errors.
Therefore, we propose to segment the trajectory by using the critical points as the segment boundaries and insert corrective motor primitives to ensure the subsequent segment's preconditions are met. Fig.~\ref{fig:methodimplementation} illustrates a trajectory with two such segments, encoded with two movement primitives, namely Pre-Movement Primitive (\Initial{}) and Post-Movement Primitive (\Final{}), and a single critical point where a corrective movement primitive (\Correction{}) enables the robot to recover from different failure situations.
In this work, we assume the critical points are known and focus on \Correction{} learning through human demonstration.
Since the correction would depend on the robot and environment configuration, i.e. the context reached after the execution of the preceding segment, we learn a context-dependent correction policy. In plain terms, a context is a vector extracted from sensors available to the robot that captures the task-relevant environment-robot relations.   
  By imposing such a task decomposition, rather than learning a monolithic task policy from demonstration, we aim to improve performance and robustness, while achieving a general correction policy for the critical point under consideration. To achieve a similar performance from a monolithic behavior cloning system, one would need to ask for multiple demonstrations for the full action; whereas in our proposal, the corrections for the critical points have to be demonstrated multiple times while the main task execution can be demonstrated a few times (typically once suffices). 

\subsection{General Framework}

Our model is built on top of Conditional Neural Movement Primitives (CNMPs) architecture \cite{Ugur-RSS-19}.  CNMPs is a deep neural architecture capable of representing sensorimotor data conditioned on the given set of observation points as a latent vector, which then can be used to unfold the trajectory based on queried time points. Thus, when coupled with a feedback controller, it can act as a drive for the movements of a robot. In our work, for the critical segment learning, we feed CNMPs with the context information so that the correction is conditioned on the context. A CNMPs is composed of two networks: an encoder and a decoder, as shown in Fig.~\ref{fig:cnmpwithcontextgraph}. The encoder part learns to generate a latent representation (i.e. latent vector) from the samples taken from an input trajectory. By taking the mean of the representation for each sample, a general representation is obtained that is sufficient to reconstruct the trajectories learned. Since training and querying are done with a context input appended to the latent representation (Fig.~\ref{fig:cnmpwithcontextgraph}), the generated trajectories become context-dependent. Thus, for the corrective motor primitives (cor-MP), CNMPs learn to generate different motor patterns based on the context. Priming the CNMPs with a context not only enforces the selection of the right correction policy but also allows generalization to unseen contexts, thereby bringing robustness to overall task execution.




In this study, we test our method on two tasks realized in simulated and real robots. Both tasks involve a single critical region. Therefore, we can unambiguously refer to pre-critical motor primitive (\Initial{}), corrective motor primitive (\Correction{}), and post-critical motor primitive(\Final{}) in our descriptions (see Fig~\ref{fig:methodimplementation}). 
Further details are explained comprehensively in Section \ref{experiments}. 

For the simulation task, human demonstration is emulated through a deterministic algorithm. On the other hand, for the hardware task, human demonstration is obtained through a teleoperation system that is developed by using two haptic devices (GeoMagic Inc.).




\subsection{Learning the Context}
The context vector provides information about the current robot state and environment relation. As such, it can be a manually designed feature extractor, or it can be learned in an end-to-end fashion through an auto-encoder (AE). To test the latter approach, for the simulation experiment we designed an AE with 2D convolutional layers of kernel sizes 3, strides [2, 1, 2, 1, 2], and filter sizes [32, 32, 64, 64, 64] in the encoder block. The output of the encoder is flattened and fed into a fully connected layer to get the latent representation of the image. The decoder mimics the encoders operations in reverse and used only in training the AE. In the simulation experiments, the encoder receives an image and the resulting latent vector is used as the context.
The AE is trained on a data set consisting of 400 images. 360 of them are used for training and the remaining 40 for testing. The images are taken after the first push and correction actions (described in Section \ref{simulation_exp_desc}). The AE is trained for 70 epochs using Adam optimizer \cite{kingma2014adam} with a learning rate of 3e-4. The size of the latent space representation of images was set to 8 for the experiments reported in the current paper.

\begin{figure}
\centering
\includegraphics[width=0.25\textwidth]{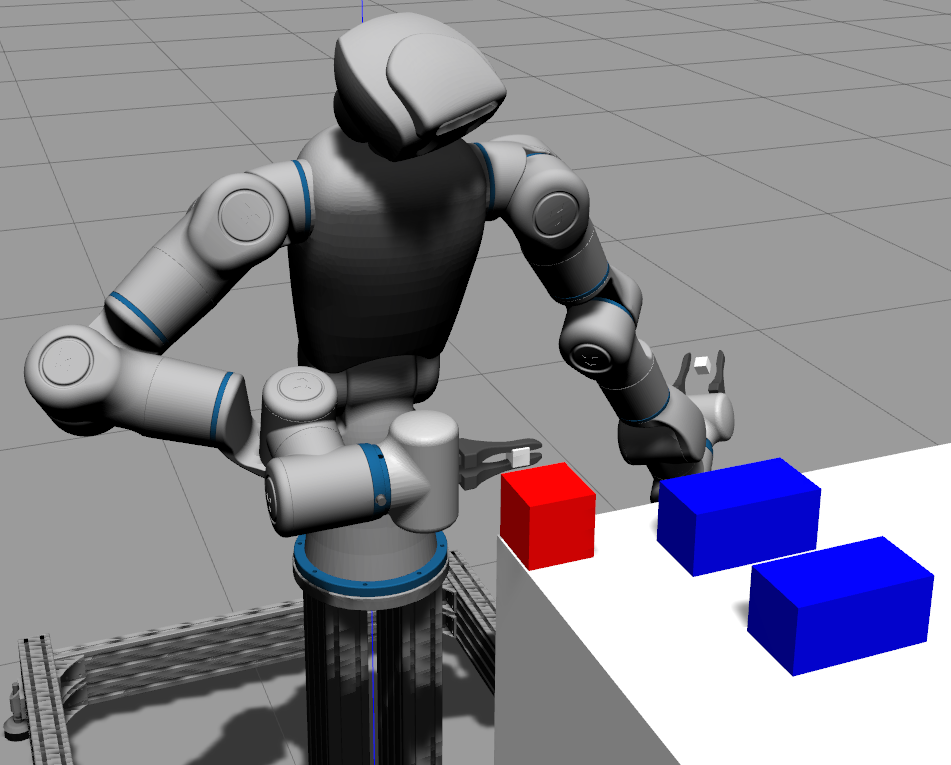}
\caption{\label{fig:simulation_initial_position} Simulation environment prepared in Gazebo.}
\vspace{-5mm}
\end{figure}


\begin{figure}[t]
    
\centering
\includegraphics[width=0.4\textwidth]{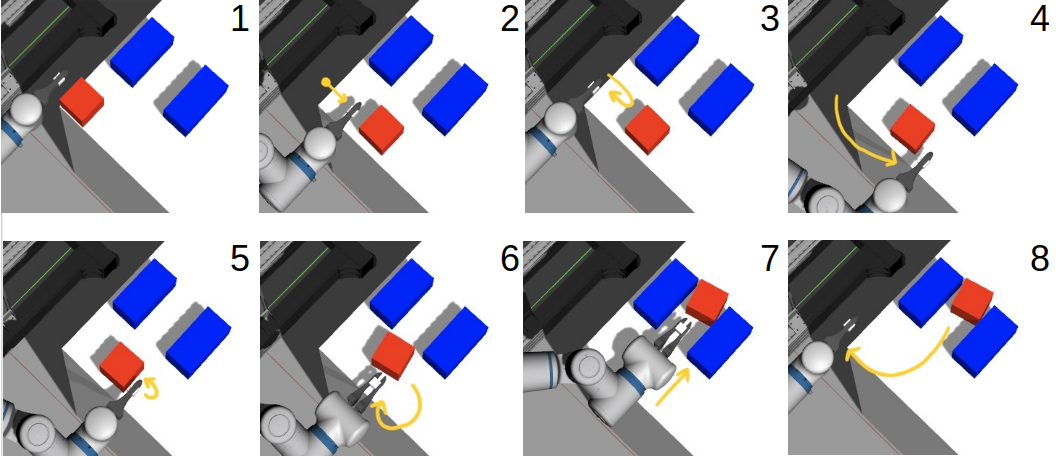}
\caption{\label{fig:sample-trajectory} A typical action execution to accomplish the goal.}
  
\end{figure}





\label{realropemanipulation}

\section{Experiments and Results}
\label{experiments}
\subsection{Push object through corridor with AE-based context}

 \label{simulation_exp_desc}
We use the simulated upper body humanoid robot Torobo (Tokyo Robotics Inc.) in Gazebo simulation environment (Fig. \ref{fig:simulation_initial_position}) in order to teach the robot to push a target object through a corridor in two steps: First the robot needs to align the object with the corridor, and then push it trough, which is a simple task the robot can perform. However, to test our method, we, perturb the position of the corridor stochastically in a 6 cm range, thereby deliberately creating a critical region for the task execution that would require a corrective step. 
Following the logic of the proposed model, the robot asks demonstrations for the those critical points that are encountered after the first push. 
As stated in Section \ref{methods}, the human demonstration is emulated by an algorithmic logic that guides the robot to do the 'right' corrective action. While the correction is being carried out, the data collected as the corrective action demonstration data, and used for training \Correction{} neural network.

The characteristics of the corrective motor primitives (\Correction{}) differ according to the position of the  corridor. When the corridor moves away from the robot, the robot needs to push the object to align it with the corridor. On the other hand, if the corridor comes closer, the robot needs to pull the object so as to align it with the corridor. A typical task execution with a pull type  correction is shown in Fig. \ref{fig:sample-trajectory}.

\begin{figure}[t]
    \centering
    \subfloat{\includegraphics[width=2cm]{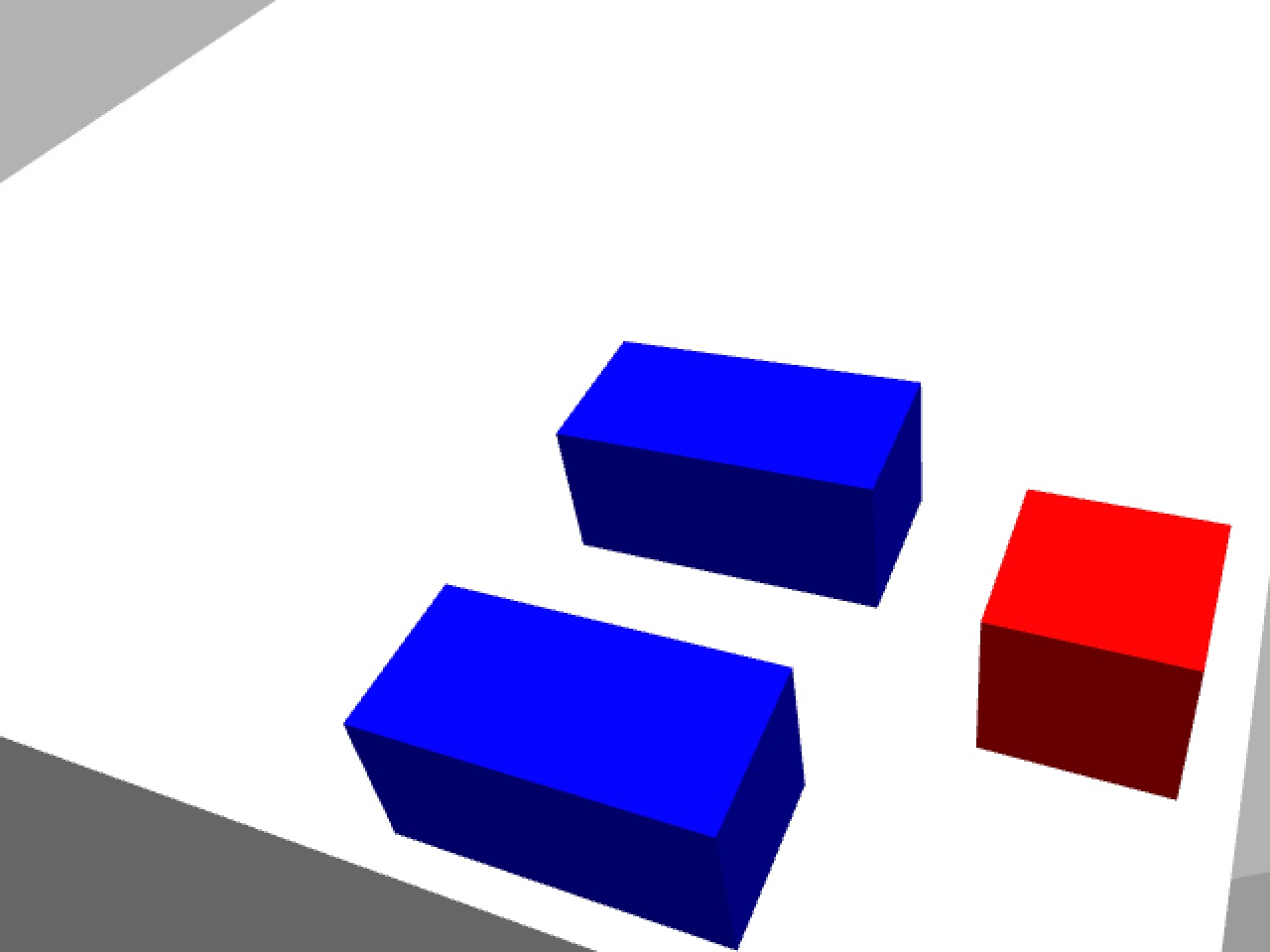} }%
    \quad
    \subfloat{\includegraphics[width=2cm]{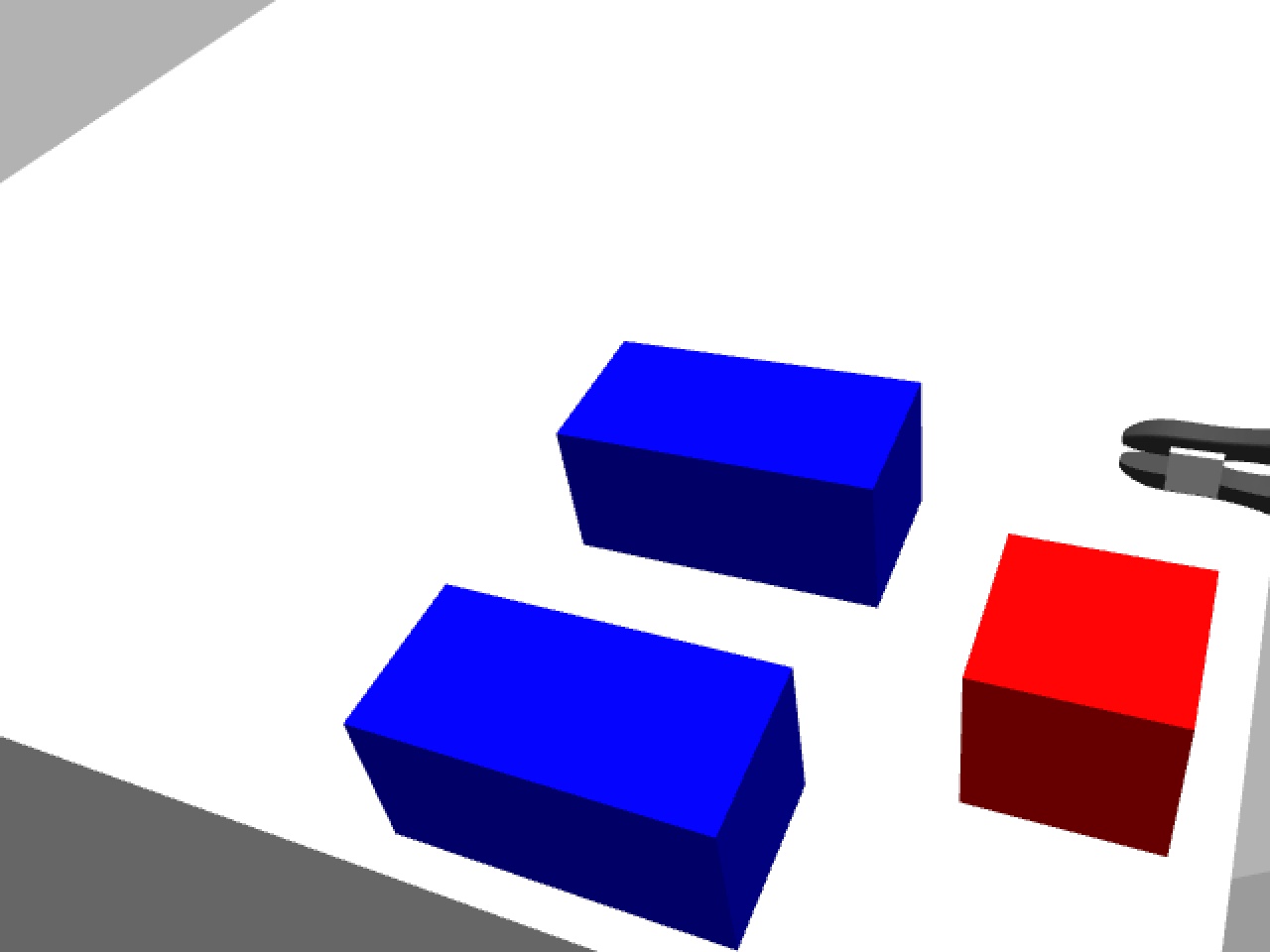} }%

    \caption{ Camera images of the scene, showing the configurations before (left) and after (right) the correction action.}%
    \label{fig:camera-images}%
    
    \vspace{-5mm}
\end{figure}

Two hundred trajectories are collected together with environment information in total to create a data set. Each data unit consists of the Cartesian position of the gripper, the corridor location, and the target object. In addition, for each trajectory, two images are captured from the camera of the Torobo Robot. The first image is captured after the execution of the pre-critical movement primitive (\Initial{}). The second image is captured after the execution of \Correction{}. These image are used to obtain autoencoder based context for the experiments that involve learned contexts (Section IV.B.2). Typical images captured from the camera of the robot head camera is shown in Fig. \ref{fig:camera-images}.

\begin{table}[b!]
  \begin{center}
    \centering
    \caption{Simulation Training Parameters}
    
    \label{tab:simtrainingtable}
    \begin{tabular}{p{1.4cm}|p{1cm}|p{1.6cm}|p{1.6cm}|p{1cm}}
      \textbf{Network Type} & \textbf{Learning Rates} & \textbf{Encoder Layer Count}  & \textbf{Decoder Layer Count} & \textbf{Layer Size}\\ 
      \hline
      All CNMPs & 0.0001 & 3 & 4 & 128\\ 
    
    \end{tabular}
  \end{center}
  \vspace{-5mm}
\end{table}

As mentioned in Section \ref{methods}, the task execution is composed of the executions of three sequential motor  primitives. Thus our system is required to produce three trajectories,  namely \InitialCNMP{}, \CorrectionCNMP{}, \FinalCNMP{} to successfully complete the task. The trajectories are encoded in task space in this experiment.
Particularly, the context is used by the \FinalCNMP{}  to make the final push since after \CorrectionCNMP{} correction, the location of the object would be, in general,  changed.
Two simulation experiments are conducted, one with 'perfect context', and one with the 'AE-based' context. 
For the perfect context, the corridor location and the object location after the correction are used.
To be concrete, \CorrectionCNMP{} uses the position of the corridor on the posterior-anterior axis as the context, and \FinalCNMP{} uses the position of the target object on the same axis. The context values are normalized before being used in \CorrectionCNMP{} and \FinalCNMP{}.

For the AE-based context based experiments, the captured images from the camera are fed into an autoencoder (AE) to extract the information related to the target object and the corridor positions. Then, the encoded images (i.e latent vectors) are given as contexts to the \CorrectionCNMP{} and \FinalCNMP{}, respectively. Table \ref{tab:simtrainingtable} shows the learning parameters used for training. To reiterate, the CNMP networks are trained by using (normalized) time and  x, y, and z positions of the gripper collected during demonstrations for both perfect and AE-based context experiments. The CNMP weights are trained by using Adam optimizer\cite{kingma2014adam}.
During the training, 100 of 200 trajectories are used for training our model for each context type. The remaining ones are used for validation. Each model received 10M weight updates. 


\subsection{Results of Push through corridor task Experiment (Simulation) Experiments}


\begin{table*}
  \begin{center}
    \centering
    \caption{The action decision and task success of our proposed architecture with perfect context (PC) and autoencoder features (AE) are provided. For comparison, the results obtained from monolithic networks (Mono) is also provided. }
    \label{tab:cartesian_pred_results}
    \begin{tabular}{|c|c|c|c|c|c|c|c|c|c|c|}
      \hline
      \multirow{2}{0.1\columnwidth}{Expected Correction} &\multicolumn{3}{|c|}{Action Choice} & \multicolumn{3}{|c|}{Decision Match} & \multicolumn{3}{|c|}{Task Success}  \\ 

      \cline{2-10}
      &PC & AE & Mono & PC & AE & Mono & PC & AE & Mono\\
      \hline
      Push & 9 Push & 8 Push, 1 No Correction & 9 Push & \( 100 \% \)  & \( 89 \% \)  & \( 100 \% \) & \( 100 \% \)  & \( 100 \% \) & \( 44 \% \) \\ 
      No Correction &  6 No Correction & 6 No Correction  &  6 No Correction& \( 100 \% \)  & \( 100 \% \)  & \( 100 \% \) & \( 83 \% \)  & \( 100 \% \) & \( 67 \% \)\\ 
      Pull &  10 Pull & 9 Pull, 1 No Correction &  10 Pull &  \( 100 \% \)  & \( 90 \% \)  & \( 100 \% \) & \( 100 \% \)  & \( 100 \% \) & \( 90 \% \)\\ 
      \hline
    \end{tabular}
  \end{center}
  \vspace{-2.5mm}
\end{table*}

\begin{table}
  \begin{center}
    \centering
    \caption{Prediction Errors (cm) of MP trajectories. }
    \label{tab:cnmp_compare_trajectory}
    \begin{tabular}{|c|c|c|c|c|c|c|c|c|c|c|}
      \hline
      \multirow{2}{0.1\columnwidth}{Network Type} & \multicolumn{2}{|c|}{\Correction{}} & \multicolumn{2}{|c|}{\Final{}}  \\

      \cline{2-5}
       & mean  & max  & mean  & max \\
      
    \hline
      PC   & 0.80 & 1.45    & 1.69  & 3.19  \\ 
      
     \hline
      AE  & 0.80 & 0.99   &  2.64 &  4.83  \\ 
      
      \hline
      B  & 1.30 & 1.92   &  5.75 &  7.24  \\ 

      \hline
    \end{tabular}
  \end{center}
  \vspace{-5mm}
\end{table}

The trained pre-MP, corr-MP and post-MP networks are integrated into the simulation environment to assess the performance of the proposed system in driving the Torobo Robot for successfully pushing the target object through the corridor. For execution, the predicted \Initial{} Cartesian trajectory is mapped to a joint trajectory through inverse kinematics which is fed to the robot system as desired joint angles at a nominal rate. The robot controller then tracks the obtained joint trajectory and ends up in the critical region. The robot observes the scene, obtains the latent vector from the trained auto-encoder (in the case of AE based context) or logs the position of the corridor (in the perfect-context case), and uses it as the context for the \CorrectionCNMP{}. By taking the last gripper position as the initial observation for \CorrectionCNMP{}, cor-MP trajectory is predicted. 
Since the \CorrectionCNMP{} has learned corrections from demonstrations, the generated trajectories are expected to correspond to push, pull, or no action (as these have been strategies used in the demonstrations). 
After executing \Correction{}, the \Final{} trajectory is predicted according to the observed context and the position of the end effector. After the execution, the trial is counted as a success one if the target center of the object passes the corridor's midpoint without touching the corridor's entrance. To asses the performance of the system 25 trials (test data set) with unseen contexts are executed with the predicted \Initial{}, \Correction{}, and \Final{} for both perfect context and AE-based context experiments.

\subsubsection{Results of \CNMP{} with Perfect Context (\PCCNMP{})}

In order to successfully complete the task, the robot was required to make 9 push and 10 pull correction actions in the 25 cases tested. The remaining 6 cases required no correction (as the object was aligned with the corridor to a tolerable level). 
The \PCCNMP{} results show that the network was able to predict an appropriate correction trajectory when faced with previously unseen contexts. This shows that the \Correction{} network can  learn to generate correct decisions and correct movement amounts. As a result of these corrective actions, the robot was able to move the object to a position to align it with the corridor. The only task failure was detected during a push type of correction generated by the \Correction{} network, and the failure was due the object slightly contacting the corridor at the entrance.

\begin{figure}[b]
\centering

\includegraphics[width=0.4\textwidth]{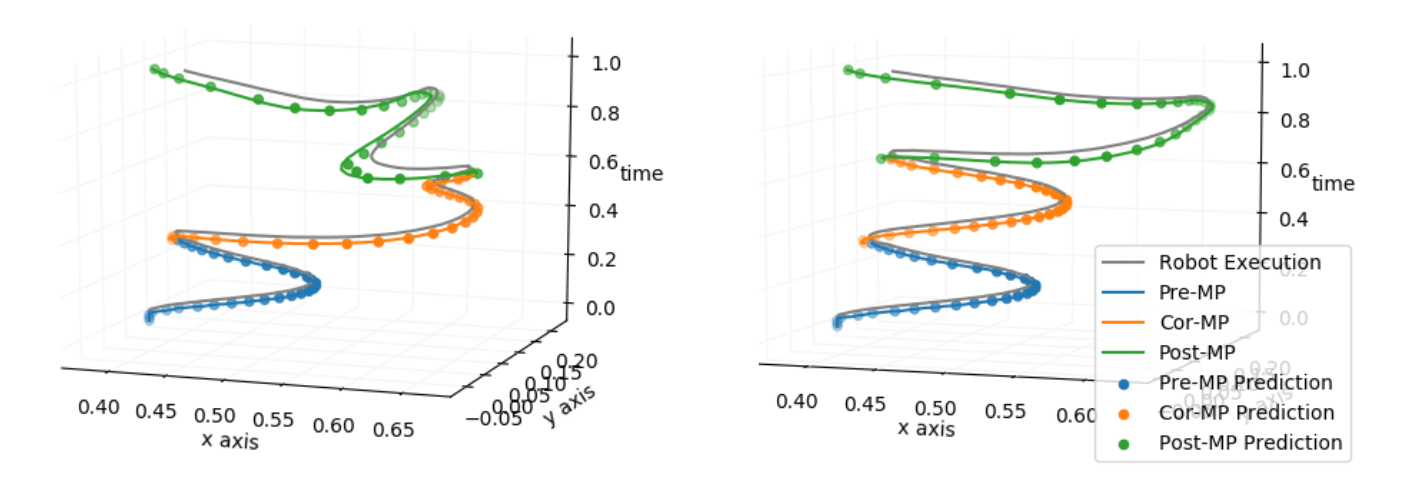}

\caption{\label{fig:gt_pred_m3} Comparison of ground-truth, prediction, and robot execution for different \Correction{}s.}
\end{figure}

\tuba{The mean and maximum error values for \Correction{} and \Final{} are shown in Table \ref{tab:cnmp_compare_trajectory}. The error for each motion is calculated by taking the average of the Euclidean distances between the predicted action points and the ground truths. The mean error is 0.80 cm for \Correction{} and 1.69 cm for \Final{}. \Final{} error is higher than the previous one because the desired motion for \Final{} is more complex than the previous one. These errors cause small noise in the end position of the target object after each motion. Because these errors are small enough, 24 trials out of 25 yielded  successful task executions. Predictions, robot executions, and the ground-truths for two different action type for \Correction{} are compared in Fig. \ref{fig:gt_pred_m3}. As seen in the Figure, the predictions overlap the ground truths. Therefore, the simulated robot was able to execute the predicted motions accordingly.}

\subsubsection{Results of CNMPs with AE (\AECNMP{})}

As shown in Table \ref{tab:cartesian_pred_results}, when AE features were used as context by our system, in 24 trials out of 25, the robot was observed to accomplish the task by pushing the target object in front of the narrow corridor, correcting its position, and pushing it through the narrow corridor to the goal position.
When the execution of each predicted motion is examined, the  \Correction{} network was mostly able to choose the right action type given the context. When we examine the predictions that perform differently from  
the test data set used to assess the performance of the predictions 
it is observed that the predicted trajectories by the \Correction{} network slightly shifted in comparison to demonstrated trajectories. 
Since the captured image at boundary cases between push and no-correction as well as pull and no-correction are similar, the AE representation for those cases are slightly smeared. Therefore, the \Correction{} network may choose to generate different action types around the decision boundaries. Still, because the correction amount that is required around the decision boundaries is small compared to the width of the corridor, the different decisions at these boundaries have not affected the success of task execution. As a result, the success of the system with AE context has been found to be 100\%.

The mean \Correction{} error is similar to that of perfect context version of the system. The mean error resulted from \Final{} network prediction is 2.64 cm, which is higher than the one found for the perfect context experiments. This might be due to the fact that the AE based context version of the system made different choices when doing the correction, leaving the target object in different location experienced during the demonstrations, which is acted upon by  \Final{}.

\begin{figure*}[ht]
\vspace*{-7mm}
\centering
\includegraphics[width=0.9\textwidth]{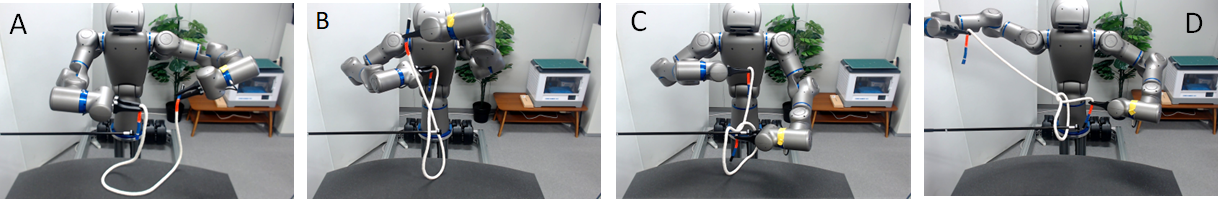}
\caption{\label{fig:knottingtherope} Knot over bar execution illustrated with Torobo robot.}
\vspace{-5mm}
\end{figure*}

\subsubsection{Comparison with Monolithic Baseline CNMP}
To assess the benefits of using segment action representation with explicit corrective motor primitives for the critical regions, we implement a monolithic baseline CNMP network. Given a fixed context, the full trajectory can be learned with the  \Baseline{}. However, since the position of the corridor (i.e. context) changes trial-to-trial, without knowing any information about the context, the learning system cannot decide which trajectory to generate. Therefore, to be fair to the baseline case, we provide the corridor location to the \Baseline{} and ask it to learn the demonstrated full trajectories (which include the proper corrections). The number of parameters of \Baseline{} is adjusted to be the sum of number of parameters of the pre-MP, cor-MP and post-MP networks
The comparison of the \Baseline{} network generated actions to those of  \PCCNMP{} and \AECNMP{} are shown in Table \ref{tab:cartesian_pred_results} with the Mono label. As shown, the \Baseline{} network can generate the right correction action; however, when executed, only 17 trials out of 25 succeed in completing the task. This is due to the errors stemming  from learning a longer trajectory, albeit the \Baseline{} network uses an equal number of parameters as of the sum of the   \PCCNMP{} and \AECNMP{} networks. The mean and max error values for the \Baseline{} network are also observed to be higher than the \PCCNMP{} and \AECNMP{} networks. The results show that the proposed method outperforms the baseline with the total number of learnable parameters equalized.
\subsection{Knotting with the real robot}
For the hardware implementation we chose bi-manual rope knotting over a bar, as it demonstrates the applicability of the proposed method for a range of complex tasks. With the implemented teleoperation setup (Section \ref{methods}), the users can control the pose of the individual grippers attached to the arms of the robot Torobo robot. With this, we obtain basic knotting demonstration on the robot by an expert demonstrator.
Learning from demonstration with the \Baseline{}
yields knotting behavior; however small variations such as how the gripper holds the rope ends, the length of the rope between the grippers, and invisible twists in the rope cause the robot to miss the critical `pass through the loop and pull the free end' (see Fig. \ref{fig:knottingtherope}) part of the knotting action. So, analogous to the simulation experiments, at the point just before the gripper would enclose (see Fig. \ref{fig:knottingtherope}-C), we collected multiple corrective demonstrations from the user to synthesize a corrective motor primitive to be inserted at this critical point with the aim of increasing the success of knotting.  The actions before and after this point i.e., (\Initial{},  \Final{})  can be learned easily by individual CNMPs. For the corrective part, we have collected 28 corrective action demonstrations which involve  posterior-to-anterior,  anterior-to-posterior, lateral corrections and their combinations. A few  demonstrations without any corrections were also included. All the demonstrations were learned with a single corrective CNMP (\Correction{} network) as in the simulations giving us a corrective motor primitive (\Correction{}). The correction was given as relative Cartesian positions with respect to the gripper position. Thus the correction could be applied in any place in the workspace of the robot. For the Cartesian to joint space transformation we used inverse kinematics with fixed gripper orientation (as attained after the execution of \Initial{}). As the context information, the \Correction{} network received the relative position of the rope end with respect to the gripper mid-point. The rope end position is extracted with a straightforward color-segmentation and clustering pipeline on the RGB-D image captured by the camera (Intel SR300) mounted on the head of Torobo. 
\begin{figure}[t]

\includegraphics[width=0.5\textwidth,right]{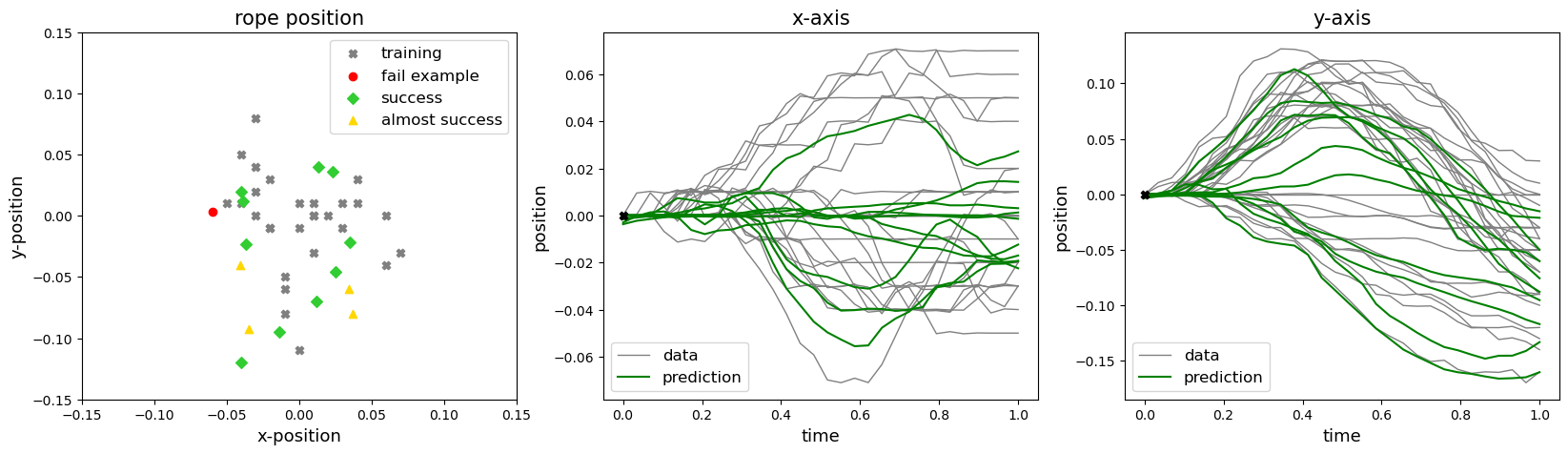}
\caption{\label{fig:datawithtestcasescontextplot} 
Real robot knotting performance. Knot correction demonstrations (gray) with predictions generated (green) and the corresponding contexts (left panel). 
}
\vspace{-6mm}
\end{figure}
The \Correction{} network could learn the demonstrations well, thus near exact context (i.e. rope-gripper configuration) at the critical point successfully generated corrective movements. More critical test was to see the performance on unseen contexts. For that, we asked the robot to execute \Initial{} and perform the correction by using the trajectory generated by the \Correction{} network, which was followed by the execution of the \Final{}, that included the enclose command of the gripper. Thus if the correction was successful, the knot operation would be completed successfully. In those trials, if the reached critical points did not require a correction, we moved the rope outside the fingers to see whether the required correction can be made. The experimental results are provided in Fig. \ref{fig:datawithtestcasescontextplot}. The gripper-robot configurations used in training are shown with gray, and the test configurations with successful correction and task execution are shown with green. In addition  some indicative failures are also  shown (yellow and red). Overall, we can conclude that the correction skill was acquired by the robot and  generalization to inexperienced contexts could be handled. Most of the failures were to due to kinematic constraints (i.e. out of reach) and others are near misses (the rope is contacted by one of the gripper fingers  but the rope was not straddled).

\section{Discussion and Future Work}


In this paper, we have proposed a novel LfD approach which can be applied to deformable object manipulation. By segmenting a given task into motor primitives and inserting explicit correction policies between the segments, we show that the potential failures that would be faced with a monolithic approach are eliminated, and accuracy and generalization of the task execution can be increased, while at the same human effort for demonstration is reduced.

In the current paper, we have focused on learning correction policies. As such the success of the proposed method relies on the knowledge on which regions are likely to need correction, i.e. critical for task success. Finding these regions automatically without excessive number of full task demonstrations would make the proposed method much more potent and applicable to tasks where domain knowledge is not readily available. 


\addtolength{\textheight}{-14cm}   



%




\bibliographystyle{IEEEtran}
\bibliography{references}

\end{document}